# Multi-dimensional data refining strategy for effective fine-tuning LLMs

**Thanh Nguyen Ngoc[1], Quang Nhat Tran[2], Arthur Tang[3], Bao Nguyen[4], Thuy Nguyen[5], Thanh Pham[6]**

**Abstract**: Data is a cornerstone for fine-tuning large language models, yet acquiring suitable data remains challenging. Challenges encompassed data scarcity, linguistic diversity, and domain-specific content. This paper presents lessons learned while crawling and refining data tailored for fine-tuning Vietnamese language models. Crafting such a dataset, while accounting for linguistic intricacies and striking a balance between inclusivity and accuracy, demands meticulous planning. Our paper presents a multidimensional strategy including leveraging existing datasets in the English language and developing customized data-crawling scripts with the assistance of generative AI tools. A fine-tuned LLM model for the Vietnamese language, which was produced using resultant datasets, demonstrated good performance while generating Vietnamese news articles from prompts. The study offers practical solutions and guidance for future fine-tuning models in languages like Vietnamese.

**Additional Keywords and Phrases:** data crawling, data refining, AI-assisted script, fine-tuning LLMs, LLMs for Vietnamese

---

[1] School of Science, Engineering, and Technology, RMIT University Vietnam
e-mail: thanh.nguyenngoc@rmit.edu.vn

[2] School of Science, Engineering, and Technology, RMIT University Vietnam
e-mail: quang.tran26@rmit.edu.vn

[3] School of Science, Engineering, and Technology, RMIT University Vietnam
e-mail: arthur.tang@rmit.edu.vn

[4] School of Science, Engineering, and Technology, RMIT University Vietnam
e-mail: bao.nguyenthien@rmit.edu.vn

[5] School of Science, Engineering, and Technology, RMIT University Vietnam
e-mail: thuy.nguyen43@rmit.edu.vn

[6] School of Science, Engineering, and Technology, RMIT University Vietnam
e-mail: thanh.pham@rmit.edu.vn

## 1 INTRODUCTION

The development of Large Language Models (LLMs) has significantly advanced the field of Natural Language Processing (NLP), enabling machines to comprehend and generate human language with unprecedented accuracy and fluency [1, 2]. LLMs demonstrated remarkable proficiency in various NLP tasks, ranging from text generation to translation and sentiment analysis. However, achieving such performance hinges on effective fine-tuning – a process where a pre-trained model is adapted to a specific task or language [3, 4].

Fine-tuning critically relies on high-quality, relevant, and representative data [5]. The significance of data in this context cannot be overstated; it plays a pivotal role in shaping a model's ability to comprehend nuances, contextual information, and idiomatic expressions specific to a particular language [6]. Additionally, fine-tuning data must accurately mirror the linguistic intricacies and cultural nuances of the target language [7]. For languages with unique linguistic characteristics and cultural contexts [8], such as Vietnamese, acquiring suitable data becomes a formidable challenge. Being a tonal language rich in diacritics and idiomatic expressions, it makes the collection and curation of data in Vietnamese even more critical due to its distinctiveness from languages like English[9].

This paper embarks on an exploration of the intricacies involved in collecting and curating data for fine-tuning Vietnamese language models. Through a systematic analysis of the challenges encountered, the paper then proposes a multi-dimensional strategy to overcome them. The work aims to provide insights that contribute to the broader discourse on data preparation for language model enhancement. Additionally, this paper aims to shed light on the complexities of data acquisition and curation in the context of refining language models for languages with distinct linguistic and cultural traits like Vietnamese.

## 2 LITERATURE REVIEW

Foundational LLMs and their fine-tuned counterparts have become a cornerstone of NLP research in recent years [10, 11]. Extensive literature has addressed the significance of data in shaping the performance of language models across various languages and tasks. Devlin et al. [12] pioneered the concept of transfer learning in NLP, showcasing the effectiveness of fine-tuning large-scale language models for specific tasks. However, while these models have demonstrated impressive capabilities across a range of languages, there is growing recognition that each language presents unique challenges that demand tailored data collection and curation approaches [13]

While a plethora of studies have underscored the importance of data quality and diversity, there remains a dearth of research specifically focusing on the challenges and strategies associated with collecting and curating data for fine-tuning Vietnamese language models [14]. For languages like Vietnamese, characterized by tonal inflections, complex diacritics, and rich idiomatic expressions, data collection becomes particularly intricate. Few studies, such as Nguyen and Le [15] have looked into language-specific challenges in NLP for Vietnamese, highlighting the necessity of linguistic expertise in dataset preparation. Despite these insights, a comprehensive exploration of the complexities, strategies, and practical solutions specific to fine-tuning Vietnamese language models remains limited.

This paper aims to bridge this gap by offering an in-depth examination of the challenges encountered during data collection and curation for fine-tuning Vietnamese language models [16, 17]. By presenting lessons learned from the process, this study contributes valuable insights that can guide researchers and practitioners in effectively harnessing data to optimize model performance for languages with unique linguistic traits like Vietnamese.

## 3 RESEARCH METHODOLOGY

Popular and commercial LLMs such as ChatGPT give unsatisfactory results for some queries in low-resource languages as their training datasets were dominantly in English. Fine-tuning LLMs would be an appropriate approach to invest in bridging the digital divide gap and increasing the inclusion of non-English speaking. As the authors of this study are living in Vietnam, the language of the study is selected to be Vietnamese. However, the presented approach can be used for other local languages such as Chinese, Hindi, Arabian, etc.

The initial target data source was a Wikipedia site in Vietnamese [18]. After 20 years of building, the Vietnamese site Wikipedia had nearly 1 million pages which would be a great source for fine-tuning. However, the contents of the site are redirected to links but most of them are pointing to empty pages, which results in many empty entries in the resultant files. More notably, Wikipedia imposes a rate limit of 1 request per second, which if violated can result in an IP address block. This restriction prevents effective data crawling as the process can take weeks. Alternatively, there have been published links for raw texts from Wikipedia, but these sources are not trustworthy and hard to curate. Another vast source considered was e-books, but this was not proceeded due to copyright concerns and cost.

There have been popular datasets used by many researchers for fine-tuning such as *Alpaca 52k* [19, 20] and *Dolly 12k* [21]. The former with 52 thousand questions and answers (and sized at 40MB), is anticipated to provide a superior fine-tuning capability than the latter. Nevertheless, these two original datasets exclusively support the English language, and hence cannot be used right away. After consideration, we selected Alpaca 52k as the foundational fine-tuning dataset and then translated it from English into Vietnamese to suit our purpose in this study.

Translating the dataset whose size is 40MB was a challenge itself. Translation services such as Google Translate, or Microsoft do not support large-size input files. Hence, we developed a Python program to split the original file into 100 smaller files and then fed them into Microsoft Word for translation. A human volunteer helped with the process. The translated texts are combined into a single file. During the translation process, some information was lost, resulting in only 33k instructions being retained in the final Vietnamese dataset.

Selecting a good base model among thousands of available open-source LLMs is challenging as the number of models released and their pace of release skyrocket. There are approximately 100,000 NLP models listed on HuggingFace, the well-known site for hosting AI models and datasets, at the time of this writing [22]. We performed an initial evaluation of the top foundational LLMs from HuggingFace LLM Leaderboard[22] against Vietnamese prompts, before picking the BigScience Large Open-science Open-access Multilingual Language Model (*BLOOM*) [23]. BLOOM 7B1 model [24, 25] would fit into a smaller GPU server which was more affordable to us. QLoRA [26]fined tuning approach was chosen as it supports 4-bit quantization and various new techniques that could significantly reduce the memory requirement. The fine-tuning was carried out using a cloud GPU service called Runpod [27], which was a Docker-based service, which took 7 hours to complete. The max steps were set as 5,000. The learning rate is set at 2e-4 while the optimizer 8-bit Adam Optimization is used. This checkpoint is saved and we used PEFT [6] to merge it with the base model to generate our new model which we refer to as *VN-BLOOM-7B1*.

In addition to the earlier translated dataset, the study approached another data source, Vietnamese online newspapers. Such sources could provide responses that are more informative with diverse writing styles, and contemporary content. The top three newspapers namely: Tuoi Tre[28], Thanh Nien [29], and VnExpress[30] were contenders. Eventually, we selected VnExpress.net due to its popularity and reach. We developed a Python program for data crawling with the assistance of ChatGPT. Despite the vast data available, the original data extracted from the news site (up to 17 thousand articles) was not ready for fine-tuning as it must be formatted in question-answer form. Human volunteers are borrowed to curate the raw version before saving it to the final dataset.

Next, the VN-BLOOM-7B1 model earlier was fine-tuned with the newspaper dataset using a similar approach and took the same amount of time as in the former round. The resulting model was referred to as *VN-BLOOM7B1-NEWS.* This second model is used to generate news articles in Vietnamese to validate both the usability of the model and the quality of the datasets developed in this work.

The HuggingFace links of the two models and the datasets can be found in the Appendices.

## 4 FINDINGS AND DISCUSSION

In this section, we will reflect on and discuss the results from the proposed data crawling and data refining. First, the VN-BLOOM7B1-NEWS model evaluation result is presented to validate the quality of the datasets. Then, each approach was discussed in detail. In summary, dataset preparation is not an easy process. For complex and unique languages, one must seek multiple data sources from different categories for a reliable dataset.

### 4.1 Evaluation of VN-BLOOM7B1-NEWS Model

Performance evaluation for an LLM is a critical cornerstone within the fine-tuning process. This holds especially true when dealing with LLMs, where the challenge surpasses mere accuracy to encompass the intrinsic value encapsulated within the generated text[31]. Conventional benchmarks like loss or validation scores exhibit limitations in offering comprehensive insights within such contexts. Hence, metrics that are tailored to the specific task have emerged as essential. For instance, in translation tasks, the Bilingual Evaluation Understudy (BLEU) [32], and in summarization tasks, the Recall-Oriented Understudy for Gisting Evaluation (ROUGE) [33] offer a more nuanced evaluation approach. Due to the intricacies of the Vietnamese language, the testing methodologies specifically tailored for validating fine-tuned models might not yield optimal results, given the need for sentiment evaluation. In this study, we attempted manual testing to evaluate the model by humans.

The VN-BLOOM7B1-NEWS model was asked to generate short news articles with suggesting titles as input prompts. The generated articles were then evaluated blindly by human volunteers focusing on writing styles, and general contents against similar types of articles from newspaper sites. Table 1 shows some generated articles with their prompts. Some personal information from articles was redacted.

Table 1: Generate articles by VN-BLOOM7B1-NEWS with the suggested title as prompt.

| Prompt | Generated responses from the fine-tuned model |
|---|---|
| Write an article about the opening ceremony of an airport. | Chính thức khánh thành sân bay [redacted] : Sáng nay (19/7), Thủ tướng [redacted] đã chủ trì lễ khánh thành sân bay [redacted]. Sân bay [redacted] có công suất thiết kế 50 triệu hành khách/năm, 5 triệu tấn hàng hóa/năm, là sân bay quốc tế cấp 4F theo phân cấp của Tổ chức Hàng không dân dụng quốc tế (ICAO). Sân bay [redacted] có tổng mức đầu tư hơn 50.000 tỷ đồng, được xây dựng trên diện tích 1.600 ha, trong đó có 1.200 ha đất sân bay, phần còn lại là đất dân cư. Sân bay [redacted] có 4 đường cất hạ cánh, đường lăn song song, đường lăn nối, nhà ga hành khách, nhà ga hàng hóa, hệ thống đường giao thông kết nối, hệ thống cấp điện, cấp nước, thông tin liên lạc, xử lý chất thải, hệ thống thoát nước mưa, nước thải, hệ thống xử lý khí thải, hệ thống xử lý nước mưa, nước thải, hệ thống xử |
| Write an article about the collaboration between a Vietnamese | |

| Prompt | Generated responses from the fine-tuned model |
|---|---|
| university with its counterpart in Australia. | Việt Nam hợp tác với Úc mở trường Đại học [redacted] tại Sydney : Theo thông tin từ Đại học Quốc gia Úc ( ANU), Đại học [redacted] sẽ được thành lập tại Sydney, Úc, với sự hợp tác của Đại học Quốc gia Úc và Tập đoàn [redacted]. Đại học [redacted] sẽ là trường đại học tư thục đầu tiên của Việt Nam và là trường đại học tư thục đầu tiên của Việt Nam được thành lập theo mô hình đại học quốc tế. Đại học [redacted] sẽ được thành lập với mục tiêu trở thành một trong những trường đại học hàng đầu khu vực và thế giới. Đại học [redacted] sẽ được thành lập với mục tiêu trở thành một trong những trường đại học hàng đầu khu vực và thế giới. Đại học [redacted] sẽ được thành lập với mục tiêu trở thành một trong những trường đại học hàng đầu khu vực và thế giới. Đại học [redacted] sẽ được thành lập với mục tiêu trở thành một trong những trường đại học hàng đầu khu vực và thế giới. Đại học [redacted] sẽ được thành lập với mục |
| Write a sharing article by an artist. | [redacted] thừa nhận thuộc giới LGBT : [redacted] là một trong những nghệ sĩ Việt Nam đầu tiên công khai giới tính của mình. Anh từng chia sẻ: " Tôi là người đồng tính, tôi không muốn giấu giếm điều đó. Tôi không muốn bị mọi người kỳ thị, tôi muốn mọi người hiểu và thông cảm cho tôi". [redacted] từng chia sẻ, anh từng bị gia đình phản đối, ngăn cấm. Tuy nhiên, anh vẫn quyết định công khai giới tính của mình. [redacted] từng chia sẻ, anh từng bị gia đình phản đối, ngăn cấm. Tuy nhiên, anh vẫn quyết định công khai giới tính của mình. [redacted] từng chia sẻ, anh từng bị gia đình phản đối, ngăn cấm. Tuy nhiên, anh vẫn quyết định công khai giới tính của mình. [redacted] từng chia sẻ, anh từng bị gia đình phản đối, ngăn cấm. Tuy nhiên, anh vẫn quyết định công khai giới tính của mình. [redacted] từng chia sẻ, anh từng bị gia đình phản đối, |

The resultant model can generate coherent and contextually relevant articles given prompts. The quality of these generated articles is such that they mirror human writing style, structure, and nuance. The blind test indicates the articles are indistinct with similar existing articles found on other news sites. The outcome validates the strong competence of the fine-tuned model, and in turn, proves the efficiency of the dataset acquired in this work using a multi-dimensional approach.

### 4.2 Leverage existing datasets

The performance of the fine-tuned model emphasizes that repurposing existing datasets and adapting them to local languages is a prominent strategy. The adaptation of Alpaca 52k significantly expedites the data acquisition process. Figure 1 shows some sample questions and answers from Alpaca 52k.

```
[
    {
        "question": "Tạo tiêu đề hợp lý cho một bài báo liên quan đến ngành công nghệ.",
        "answer": "'Công ty hàng đầu thế giới về trí tuệ nhân tạo mua lại công ty điện toán lượng tử đột phá'"
    },
    {
        "question": "Tạo ra một trò đùa ban đầu bằng cách sử dụng các từ sau: nữ tu, pizza và giường",
        "answer": "Tại sao nữ tu lại gọi một chiếc bánh pizza đến giường của mình? \n\nBởi vì cô ấy muốn thề sến sẩm!"
    },
    {
        "question": "Đưa ra hai ví dụ về các từ có chứa chữ cái ‘z’",
        "answer": "Hai ví dụ về các từ có chứa chữ 'z' là 'zebra' và 'pizza'."
    },
    {
        "question": "answer question của khách hàng: 'Chính sách hoàn trả đối với sản phẩm này là gì?'",
        "answer": "Chính sách hoàn trả cho sản phẩm sẽ phụ thuộc vào nhà bán lẻ hoặc nhà sản xuất mà sản phẩm được mua, cũng như các điều khoản và điều kiện cụ
thể được áp dụng. Bạn có thể vui lòng cung cấp cho tôi thêm thông tin về địa điểm và thời điểm mua sản phẩm để tôi có thể cung cấp cho bạn phản hồi chính xác về
chính sách hoàn trả áp dụng cho nó không?"
    },
    {
        "question": "Đối thoại với người quản lý đang yêu cầu cập nhật về tình trạng của dự án - Người quản lý: 'Tiến độ của dự án là gì?'",
        "answer": "AI: 'Theo thông tin mới nhất, dự án hiện đang đúng tiến độ với 80% hoàn thành. Nhóm nghiên cứu hiện đang làm việc trên các giai đoạn cuối cùng
và dự kiến mọi thứ sẽ được hoàn thành trước thời hạn đã định. Bạn có muốn phân tích chi tiết hơn về tiến trình không?'"
    },
    {
        "question": "Chuyển đổi số này thành chữ số La Mã: 23",
        "answer": "Biểu diễn chữ số La Mã của số 23 là XXIII."
    },
```

Figure 1: Sample translated Alpaca dataset in Vietnamese.

In this work, the integration of localized datasets such as Alpaca 52k underscores the significance of data acquisition strategies that account for linguistic diversity. While translation efforts can be formidable, the optimization of resource allocation through segmentation and reliance on available translation tools offers a pragmatic way forward. By addressing the challenges posed by language translation with innovative solutions, the efficacy of fine-tuning for language models can be further elevated, amplifying their ability to comprehend and generate content in diverse linguistic contexts. This strategy not only optimizes resource allocation but also mitigates potential bottlenecks. By adopting this approach, the balance between effort and output can be carefully calibrated, contributing to a comprehensive, diverse, and linguistically enriched dataset for fine-tuning. Future work includes the consideration of other datasets such as Dolly 12k [21].

### 4.3 AI-assisted data crawling

Data crawling from websites is another important strategy and has been used by many organizations including OpenAI for their ChatGPT [34]. Developing an efficient data crawling tool presents a considerable challenge due to the inherent variability in website structures and access limitations. Each website comes with its distinct layout and constraints, making a universal approach elusive. For instance, while initially seeming promising, attempts to extract data from sources like Wikipedia encountered stumbling blocks due to their rate limits

In light of this complexity, the use of AI tools in developing data crawling scripts, such as ChatGPT, emerges as a valuable solution to tailor code for diverse website sources in a dependable manner. The flexibility of AI-driven systems lends itself to adapting and refining the crawling process to accommodate varied web layouts. In our specific undertaking, the utilization of ChatGPT proved invaluable in guiding iterative adjustments of the code until an effective version for our use case was achieved. For instance, Figure 2 shows a script snippet developed with the assistance of ChatGPT in this work for the website VnExpress.net. The code is to extract all links from the site. Links to the ChatGPT prompts and the script are listed in Appendices.

```
import requests
from bs4 import BeautifulSoup
from urllib.parse import urljoin
import json

def get_links_from_url(url):
    try:
        if "https" in url and "vnexpress.net" in url:
            response = requests.get(url)
            response.raise_for_status()  # Check if the request was successful
            soup = BeautifulSoup(response.content, 'html.parser')
            links = []
            for link in soup.find_all('a', href=True):
                absolute_url = urljoin(url, link['href'])  # Convert relative URL to absolute URL
                links.append(absolute_url)
            return links
    except requests.exceptions.RequestException as e:
        print("Error:", e)
        return []
```

Figure 2: Python script generated by ChatGPT with our instructions for crawling all links from the VnExpress.net site.

Then, by employing ChatGPT in successive rounds of prompting and refining, we were able to fine-tune the data crawling process, molding it to match the intricacies of our target sources. The dynamic nature of AI-adapted code allows for a more adaptive and responsive approach, transcending the constraints that plagued earlier attempts. Ultimately, leveraging AI tools facilitates the creation of a more robust and versatile data crawling framework, enhancing the efficiency and reliability of data acquisition from diverse web platforms.

## 5 CONCLUSION

In conclusion, the pivotal role of data in refining large language models cannot be overstated, yet the intricate process of acquiring appropriate data continues to present a formidable challenge. Through the lens of amassing and refining data specifically tailored for fine-tuning Vietnamese language models, this paper has illuminated a spectrum of insights derived from navigating the intricacies of data collection and curation.

Practically, the paper presented two main strategies: repurposing existing and well-known datasets from English and translating them into Vietnamese, and data crawling from local sources by scripts developed with assistance from AI tools like ChatGPT. The former requires human curation on the translated dataset before it can be used for fine-tuning. On the other hand, the second approach proves to be efficient due to the increasing power of Generative AI tools in generating code. This is extremely important as data crawling for various websites varies significantly due to their intrinsic architectures.

The resultant datasets from the work were used to fine-tune a base LLM model. The outcome was the model that could generate human-like articles from given prompts. The generated articles were validated by human volunteers and demonstrated good quality, which further validated the data collection strategies proposed earlier.

In summary, by employing multidimensional data approaches, the shared lessons underscored in this paper serve to illuminate the nuanced challenges inherent in dataset preparation for the enhancement of language models. With the burgeoning progress in natural language processing, these insights are poised to play a pivotal role in the ongoing refinement and advancement of models across diverse linguistic contexts. In essence, this paper contributes not only to the broader understanding of data curation but also to the ongoing evolution of language models themselves.